\documentclass[conference]{IEEEtran}
\IEEEoverridecommandlockouts

\usepackage{graphicx}
\usepackage{amsmath}
\usepackage{amssymb}
\usepackage{booktabs}
\usepackage{multirow}

\usepackage[pagebackref,breaklinks,colorlinks]{hyperref}

\usepackage[capitalize]{cleveref}
\crefname{section}{Sec.}{Secs.}
\Crefname{section}{Section}{Sections}
\Crefname{table}{Table}{Tables}
\crefname{table}{Tab.}{Tabs.}

\begin{document}

\title{Temporal-Guided Spiking Neural Networks for Event-Based Human Action Recognition}

\author{\IEEEauthorblockN{Siyuan Yang\textsuperscript{\rm1*}, Shilin Lu\textsuperscript{\rm2*}\thanks{*: Equal contribution.}, Shizheng Wang\textsuperscript{\rm3}, Meng Hwa Er\textsuperscript{\rm1}, Zengwei Zheng\textsuperscript{\rm4}, Alex C. Kot\textsuperscript{\rm1}}
\IEEEauthorblockA{
\textsuperscript{\rm1} School of Electrical and Electronic Engineering, Nanyang Technological University, Singapore\\
\textsuperscript{\rm2} College of Computing and Data Science, Nanyang Technological University, Singapore\\
\textsuperscript{\rm3} Institute of Microelectronics, Chinese Academy of Sciences, China\\
\textsuperscript{\rm4} Department of Computer Science and Computing, Zhejiang University City College, China\\
}
\{siyuan.yang, shilin002, emher, eackot\}@ntu.edu.sg, shizheng.wang@foxmail.com, zhengzw@zucc.edu.cn
}

\maketitle

\begin{abstract}
This paper explores the promising interplay between spiking neural networks (SNNs) and event-based cameras for privacy-preserving human action recognition (HAR). 
The unique feature of event cameras in capturing only the outlines of motion, combined with SNNs' proficiency in processing spatiotemporal data through spikes, establishes a highly synergistic compatibility for event-based HAR. 
Previous studies, however, have been limited by SNNs' ability to process long-term temporal information, essential for precise HAR.
In this paper, we introduce two novel frameworks to address this:
temporal segment-based SNN (\textit{TS-SNN}) and 3D convolutional SNN (\textit{3D-SNN}). 
The \textit{TS-SNN} extracts long-term temporal information by dividing actions into shorter segments, while the \textit{3D-SNN} replaces 2D spatial elements with 3D components to facilitate the transmission of temporal information.  
To promote further research in event-based HAR, we create a dataset, \textit{FallingDetection-CeleX}, collected using the high-resolution CeleX-V event camera $(1280 \times 800)$, comprising 7 distinct actions. 
Extensive experimental results show that our proposed frameworks surpass state-of-the-art SNN methods on our newly collected dataset and three other neuromorphic datasets, showcasing their effectiveness in handling long-range temporal information for event-based HAR.
\end{abstract}

\section{Introduction}
\label{sec:introduction}
Spiking neural networks (SNNs) represent the third generation~\cite{maass1997networks} of neural networks and are distinguished for their ability to perform tasks with ultra-low power consumption when deployed on dedicated neuromorphic hardware~\cite{perez2021sparse, zenke2021brain}.
{These networks} transmit spatiotemporal information between units via discrete spikes, mimicking the biological neural system. 
{Their inherent compatibility with event-based cameras is notable. These event cameras use bio-inspired sensors to asynchronously capture changes in pixel brightness, creating a stream of events that record the time, location, and intensity of brightness changes~\cite{gallego2020event}.}
Consequently, SNNs are ideally suited for integration with event cameras.

{Previous studies on SNNs~\cite{liu2020effective, orchard2015hfirst, xiao2019event,lu2024mace, 10558608} have demonstrated significant achievements in object recognition and classification.}
We posit that the integration of SNNs and event-based cameras holds exceptional promise for human action recognition (HAR)~\cite{sun2022human}.
Conventional video-based HAR models often raise privacy concerns~\cite{dave2022spact, edu2020smart, wu2020privacy, 10558208, 10181797, lu2024robust}, 
making them unsuitable for sensitive environments, such as fall detection in private areas like bathrooms. 
The collaboration of SNNs and event cameras addresses this concern effectively. 
Event cameras capture only the dynamics of moving subjects without revealing identifiable features or static backgrounds, thus ensuring privacy \textcolor{black}{preservation} in HAR.

Recent research~\cite{liu2021event, fang2021incorporating, fang2021deep, 10181564, lu2023tf, li2025set} has shown SNNs' effectiveness for HAR on neuromorphic datasets captured by event cameras. 
{However, their handling of long-range temporal information relies exclusively on the SNNs' spiking neurons, proving inadequate for comprehensive video-based HAR.}
Effective processing of long-range temporal information is critical for accurate video-based HAR. 
In light of this, we introduce two novel frameworks to enhance SNNs' capabilities in this regard. 
The first, the temporal segment-based SNN (\textit{TS-SNN}), implements a temporal segmentation strategy~\cite{wang2018temporal} on SNNs, segmenting lengthy action sequences into shorter intervals for more efficient long-term data extraction.
The second, referred to 3D convolutional SNN (\textit{3D-SNN}), replaces 2D spatial elements in SNNs with 3D spatio-temporal components, improving temporal information flow across layers.

To advance research in privacy-conscious HAR, we develop the \textit{FallingDetection-CeleX} dataset, specifically tailored for privacy-preserving applications.
{At present, there is a scarcity of real-world, event-based HAR datasets with existing ones~\cite{amir2017low, liu2021event, miao2019neuromorphic, gao2024eraseanything} primarily focus on standard action recognition scenarios, overlooking privacy-sensitive scenarios prevalent in home environments, such as fall detection.}
Moreover, the majority of these event-based HAR datasets were recorded using DVS128 and DAVIS346 sensors~\cite{lichtsteiner2008128}, which offer limited resolutions of $128 \times 128$ and $346 \times 260$ pixels, respectively. 
In contrast, our dataset employs the high-resolution CeleX-V event camera~\cite{Chen_2019_CVPR_Workshops}, a 1-megapixel sensor with $1280 \times 800$ pixels, capturing more detailed information compared to lower-resolution alternatives. 
The \textit{FallingDetection-CeleX} dataset encompasses 875 recordings from 51 subjects performing 7 different actions, including three types of falls.

We conduct quantitative comparisons with state-of-the-art SNN methods on our {newly collected} dataset, as well as three additional datasets. The experimental results indicate that the proposed frameworks outperform state-of-the-art methods, demonstrating superior capability in handling long-range temporal information for event-based human action recognition.

The contributions of this work are threefold. 
\textit{First,} we introduce \textit{TS-SNN}, utilizing temporal segment strategies to efficiently process long-term temporal information. This approach segments lengthy actions into shorter periods, enhancing SNNs' efficiency in analyzing long-term data. 
\textit{Second,} we propose \textit{3D-SNN}, replacing 2D spatial elements in SNNs with 3D spatiotemporal components, improving the transfer and processing of temporal information in event-based HAR contexts. 
\textit{Third,} we collect an event-based HAR dataset named \textit{FallingDetection-Celex}, with a special focus on falling detection. 
The dataset includes 875 recordings of 51 subjects performing 7 distinct actions, including three types of falling.

\begin{figure}
  \centering
    \includegraphics[width=0.2\textwidth]{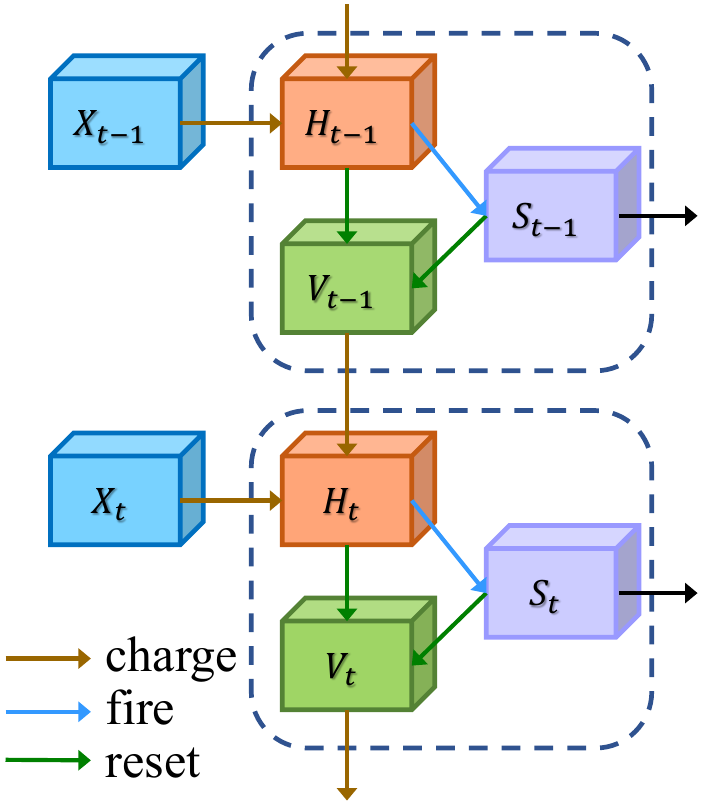}
    \vspace{-0.3cm}
    \caption{The general discrete spiking neuron model.}
\vspace{-0.6cm}
\label{fig:spiking neuron model} 
\end{figure}
\vspace{-0.1cm}
\section{Preliminary: Spiking neuron model}\label{LIF}
Spiking neuron models serve as the fundamental computational unit of SNNs. 
A unified model~\cite{fang2021incorporating} characterizes the dynamics of diverse spiking neuron types, employing the subsequent discrete-time equations:
\begin{equation}\small
	H(t) = f(V(t - 1), X(t)),
	\label{eq:5}
\end{equation}
\begin{equation}\small
	S(t) = \Theta (H(t) - {V_\text{th}}), 
	\label{eq:6}
\end{equation}
\begin{equation}\small
	V(t) = H(t) \cdot (1 - S(t)) + {V_\text{reset}} \cdot S(t),
	\label{eq:7}
\end{equation}
where $V(t)$ indicates the membrane potential after the trigger of a spike at time $t$, $H(t)$ represents the membrane potential after neuronal dynamics, $X(t)$ denotes the external input to the neuron at time $t$, $S(t)$ denotes the output spike at time $t$, and $\Theta (\cdot)$ is the Heaviside step function.
When $H(t)$ reaches a particular threshold $V_\text{th}$ at a given time $t$, the neuron generates a spike, resulting in the membrane potential to drop to a reset value $V_\text{reset}$ below $V_\text{th}$. 
This process is referred to as the hard reset and is widely used in deep SNNs. 
These equations form a general discrete spiking neuron model, illustrated in Figure~\ref{fig:spiking neuron model}.

Spiking neuron models frequently used include Hodgkin-Huxley~\cite{hodgkin1952quantitative}, Izhikevich~\cite{izhikevich2007dynamical}, and leaky integrate-and-fire (LIF)~\cite{gerstner2014neuronal} models. 
%
While these models vary in their neuronal dynamics (Eq. \ref{eq:5}), they share identical neuronal fire (Eq. \ref{eq:6}) and reset (Eq. \ref{eq:7}) mechanisms.
Among these models, 
the LIF model, being the most straightforward and efficient, is ideal for implementation. 
Its neuronal dynamics are defined by~\cite{gerstner2014neuronal}:
\begin{equation}\small
	H(t) = V(t - 1) + {1 \over \tau} \cdot (X(t) - (V(t - 1) - {V_\text{reset}})),
	\label{eq:2}
\end{equation}
where $\tau$ is the membrane time constant.
\cite{fang2021incorporating} introduces a training algorithm that enables the learning of both the synaptic weights and membrane time constants in LIF  neurons, known as parametric leaky integrate-and-fire (PLIF) neurons. In this paper, we utilize PLIF neurons as the computational units to enhance the overall expressiveness of SNNs. The neuronal dynamics is defined by Eq.~\ref{eq:learnable}, where $a$ is a learnable parameter.
\begin{equation}\small
	H(t) = V(t - 1) + {1 \over {1 + \exp ( - a)}}\left(X(t) - (V(t) - {V_\text{reset}})\right).
	\label{eq:learnable}
\end{equation}

\vspace{-0.1cm}
\section{Method}

In this section, 
we propose two novel frameworks inspired by video processing techniques: \textit{TS-SNN} (Section~\ref{sec: tsn}) and \textit{3D-SNN} (Section~\ref{sec:3d}).   
These approaches enhance SNNs' ability to handle long-range temporal information, thus improving their performance on lengthy video inputs.

\vspace{-0.15cm}
\subsection{Neuromorphic Preprocessing}
\begin{figure}
    \centering
    \includegraphics[width=0.36\textwidth]{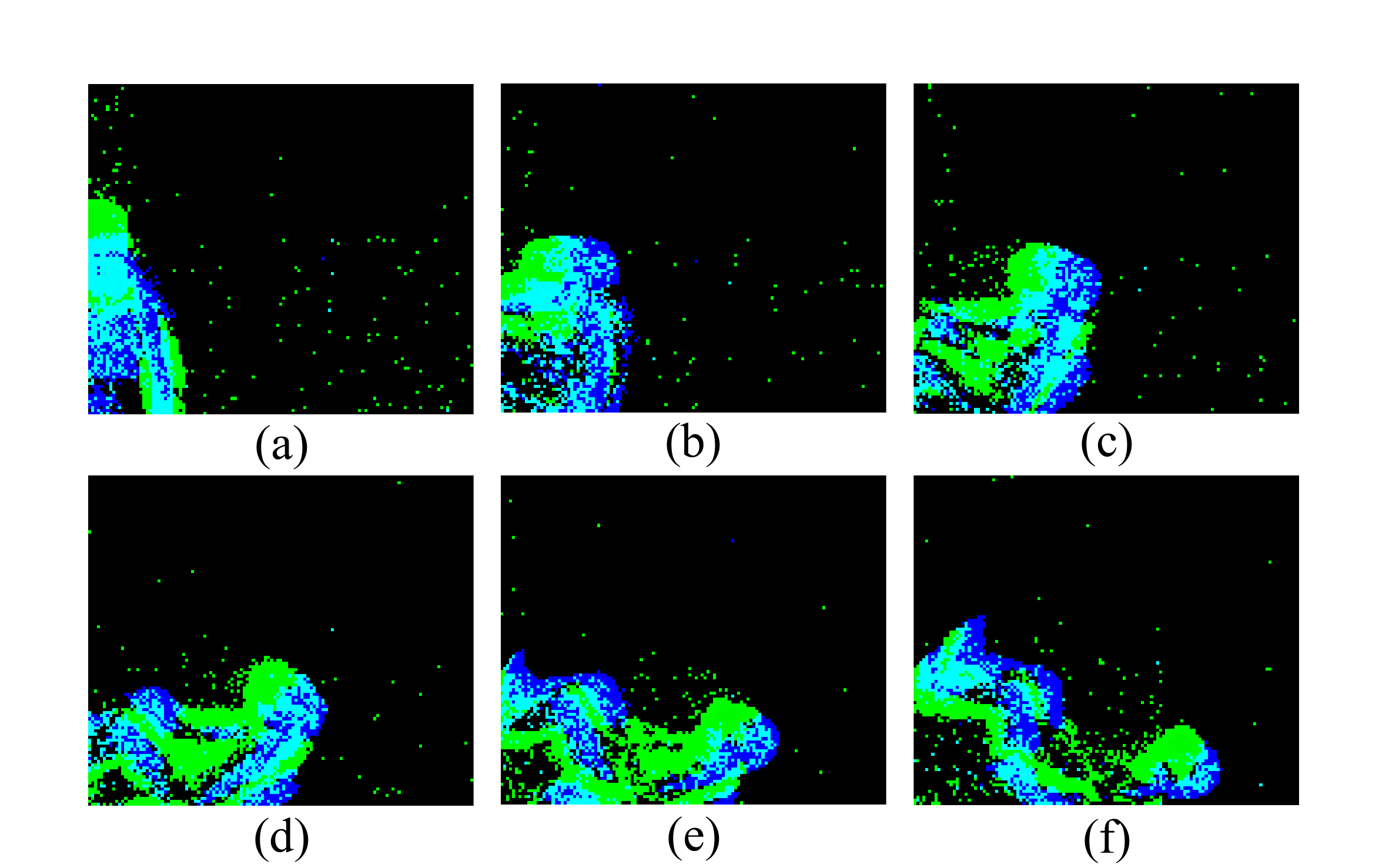}
    \vspace{-0.4cm}
    \caption{Video frames converted from event data, depicting a person falling down, captured from a side view. Frames are sequenced from (a) to (f).}
    \vspace{-0.6cm}
    
    \label{fig:Event} 
\end{figure}

Our research focuses on neuromorphic action recognition datasets from event cameras, processed using event-to-frame integration techniques~\cite{he2020comparing,neil2016phased,neil2016effective,wu2019direct,wu2022brain,xing2020new,cheng2020lisnn,kaiser2020synaptic,lee2016training} to create two-channel videos.
The event data, represented as $e({x_i},{y_i},{t_i},{p_i}) (i = 0,1, \cdots,N - 1)$, captures the pixel location of brightness changes ($x_i$ and $y_i$), timestamp ($t_i$), and polarity ($p_i$) of the event.
This data is split into $T$ slices with a similar number of events in each slice before being integrated into frames. 
The final tensor produced has the shape of $[T,2,H,W]$, where $H$ and $W$ are the frame dimensions.
%
The event data is subsequently transformed into two-channel videos, capturing only dynamic actions and excluding static backgrounds and detailed individual features, as shown in Figure~\ref{fig:Event}.
This approach effectively preserves user privacy while providing essential information about their actions.

\vspace{-0.15cm}
\subsection{Temporal Segment-Based Spiking Neural Network}
\label{sec: tsn}

Long-range temporal information is crucial for accurate human action recognition. 
A major limitation of conventional SNNs is that components other than spiking neurons struggle to model long-term temporal information effectively.
%
This is due to their limited temporal context, as they're designed to operate on single frames, typical of spatial networks.
However, complex actions like falling down involve multiple stages over extended periods. Failing to utilize long-term temporal structures in SNN training would significantly impair performance.

\begin{figure}
    \centering
    \includegraphics[width=0.4\textwidth]{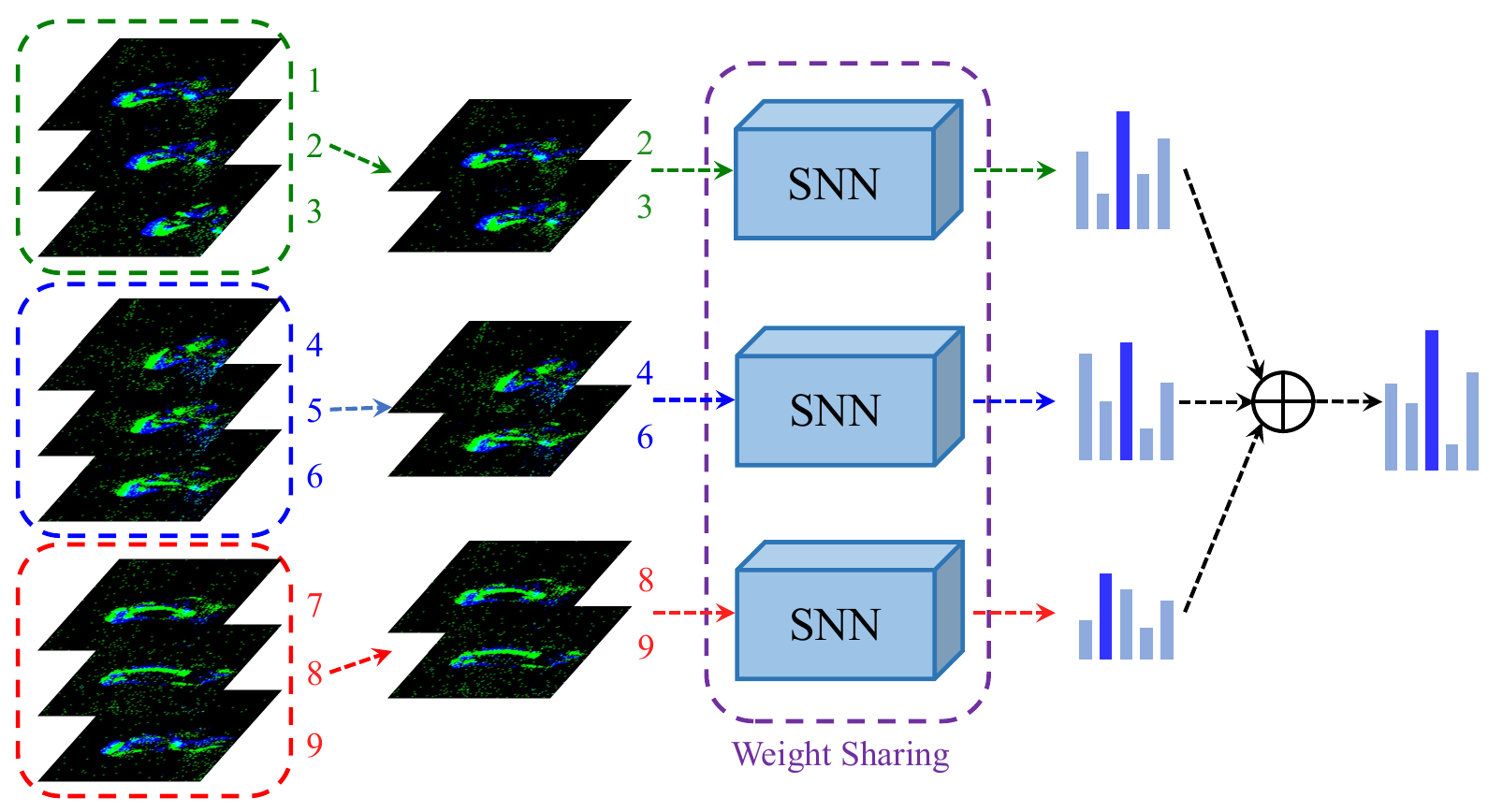}
    \vspace{-0.2cm}
    \caption{The framework of \textit{TS-SNN}. As an example, a video comprising $N=9$ frames is partitioned into $L=3$ segments, with $K=2$ frames randomly selected per segment.
    These are processed through a weight-shared SEW ResNet, and the resulting distributions are combined to make predictions.}
    \label{fig:TSN} 
   \vspace{-.5cm}
\end{figure}

As shown in Figure~\ref{fig:TSN}, we propose incorporating the temporal segment network (TSN) strategy, used in video action recognition, into SNNs.
Since the spatial components in SNNs are inherently designed to operate on a single frame, their prediction of short-term temporal information is more precise than long-term temporal information.
Consequently, we divide the preprocessed video into $L$ segments ${S_1, S_2, ..., S_L }$, segmenting the long-range temporal information into shorter segments.
From each segment, $K$ frames representing the same action moment are randomly selected and processed by a single SNN, yielding precise short-term temporal predictions.
%
These SNNs share weights across segments, producing $L$ accurate short-term prediction distributions. 
%
Finally, these distributions are then merged using simple fusion methods (e.g., summation, averaging, or maximum operation) to create a comprehensive long-term distribution. 
This technique boosts SNNs' capacity to process extended temporal information, enhancing the accuracy of event-based human action recognition.
Formally, the ultimate accurate long-term distribution is formulated as:
\begin{equation}\small
y = \text{Softmax} \left( {\upsilon (\varphi ({S_{1}};{\bf{W}}),\varphi ({S_{2}};{\bf{W}}), \cdots ,\varphi ({S_{L}};{\bf{W}}))} \right),
\end{equation}
where ${S_i}$ indicates the $i$th segment comprising $K$ frames. $\varphi ({S_i};{\bf{W}})$ is the weight-shared SNN that processes each segment $S_i$ and produces its corresponding prediction distribution. 
The segmental consensus function $\upsilon$ combines the distributions across segments to obtain a consensus of class prediction $y$ among them. 
This approach enhances the ability of SNNs to handle long-term temporal information, thereby improving the accuracy of {event-based} human action recognition. 

\vspace{-0.1cm}
\subsection{3D Spiking Neural Network}
\label{sec:3d}
\begin{figure}[t]
    \centering
    \includegraphics[width=0.25\textwidth]{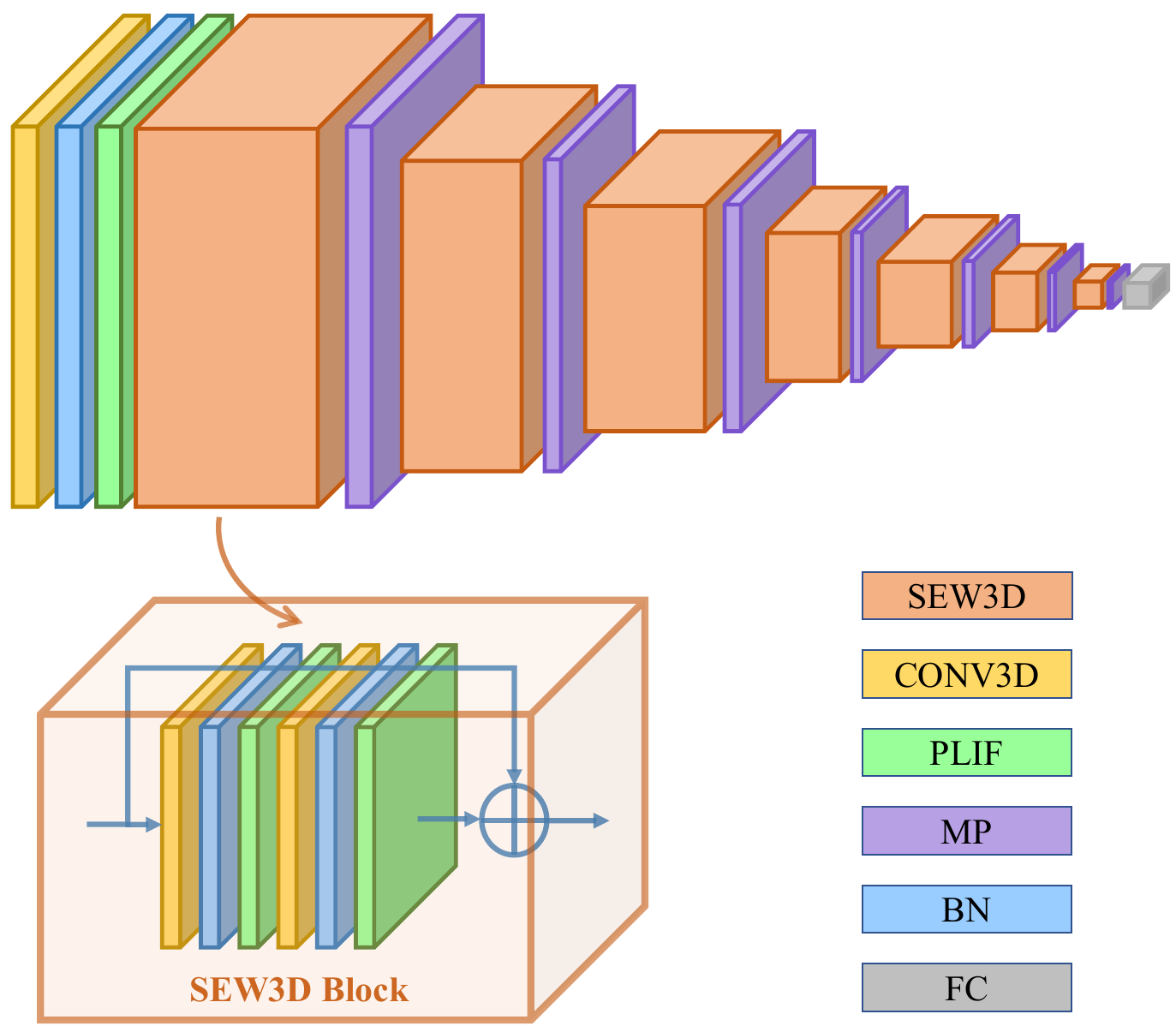}
    \vspace{-.3cm}
    \caption{Architecture of \textit{3D-SNN}, with 7 spike-element-wise (\textit{SEW}) residual blocks, each including \textit{CONV3D}, \textit{BN}, \textit{PLIF}, \textit{MaxPooling}, and \textit{FC} layers.}
    \label{fig:SEW}
   \vspace{-.5cm}
\end{figure}

Although the temporal segment strategy has significantly improved the performance of SNNs in event-based human action recognition, the spatial components of the SNN architecture still process frames sequentially, resulting in temporal information delivery occurring only in spiking neurons and the step of distribution incorporation.
To further improve SNNs' temporal information handling, we must enhance network components beyond spiking neurons. 
The inability of these components to handle temporal information limits the model's understanding of the video. 
3D convolution is a suitable candidate that effectively preserves spatial-temporal information and is compatible with spiking neurons. 
Therefore, we propose replacing the 2D spatial components with 3D {spatial-temporal} components to facilitate the delivery of temporal information between different layers. 
Our framework is illustrated in Figure~\ref{fig:SEW}. It is important to note that {SNNs} can be enhanced with 3D components to better handle temporal information.

\section{Experiments}

\vspace{-0.15cm}
\subsection{Datasets}
\noindent\textbf{FallingDetection-CeleX} Dateset: \textit{FallingDetection-CeleX} dataset comprises 875 recordings of 51 subjects acting 7 different actions, namely lying down, sitting, squatting, bending, falling from a standing position, falling while getting up and falling backward/slipping. {The CeleX-V~\cite{Chen_2019_CVPR_Workshops} event camera is used to do the recording.}
We select 581 clips for training and 294 clips for testing. In order to capture event sequences from various angles, each subject repeats each action from 3 different viewpoints (i.e, front, back, and side views). 
We consider two tasks in this dataset: the normal 7-class `Action Recognition', and the `Falling Detection', which is the binary classification problem (fall down or not). 

\noindent\textbf{DVSGesture} dataset \cite{amir2017low}: It is a real-world gesture recognition dataset collected by the DVS128 camera with a sensor size of $128 \times 128$. It comprises 1342 recordings of 11 different actions collected on 29 individuals under 3 different lighting conditions.
As suggested by the dataset paper, 23 subjects are used for training and the remaining 6 for validation.

\noindent\textbf{DailyAction} dataset \cite{liu2021event}: 
It comprises 1440 recordings of 15 subjects performing 12 different actions. 
A DVS camera was positioned at two different locations, each with a distinct distance and angle.
The actions were recorded under two lighting conditions: \textit{natural light} and \textit{LED light}.
%

\noindent\textbf{Action Recognition (AR)} dataset~\cite{miao2019neuromorphic}: It has 291 recordings of 15 subjects acting 10 different actions, captured using DAVIS cameras placed at three different distances from the subjects. Each subject performed each action three times.


Both DailyAction~\cite{liu2021event} and AR dataset~\cite{miao2019neuromorphic} mention that the datasets are split into training and test sets in an 8:2 ratio. We conduct experiments using a 5-fold cross-validation strategy.


\vspace{-0.15cm}
\subsection{Implementation Details}
\label{sec: implementation}
We select the SEW ResNet \cite{fang2021deep} as our baseline model, and train both \textit{3D-SNN} and \textit{TS-SNN} from scratch in an end-to-end manner, using cross-entropy loss as the classification loss.
The network architecture consisted of 7 blocks for all experiments. In the \textit{3D-SNN}, the first convolutional layer implemented a $1 \times 1 \times 1$ kernel, while the following 3D convolutional layers employed a $1 \times 3 \times 3$ kernel size with 32 channels for the DVSGesture dataset and a $3 \times 3 \times 3$ kernel size with 128 channels for the other three datasets. 
Each max pooling layer used a $1 \times 2 \times 2$ kernel size. 
For \textit{TS-SNN}, we follow the network setup in SEW ResNet \cite{fang2021deep} for all four datasets. 
For the DVSGesture dataset, we use Stochastic Gradient Descent (SGD) as the optimizer with an initial learning rate of 0.001, adjusted using cosine annealing. 
Both \textit{3D-SNN} and \textit{TS-SNN} were trained for 192 epochs with a batch size of 16, consistent with the baseline models \cite{fang2021deep}. 
For the other three datasets, the Adam optimizer is utilized, with the initial learning rate also set at 0.001 and adjusted by cosine annealing. The total training duration is 300 epochs with a batch size of 8.
For the \textit{3D-SNN}, all datasets are integrated into sequences of 16 frames ($T=16$), with 12 frames ($T_\text{train}=12$) randomly selected for training. 
For the \textit{TS-SNN}, all datasets are integrated into sequences of 24 frames ($T=24$); these sequences are divided into three segments ($L=3$), and five frames ($K=5$) are randomly selected from each segment.

\begin{table}[t]
\caption{Comparisons to state-of-the-art methods on four datasets. *: re-implementation results based on publicly available code. 
}
\begin{center}
\footnotesize
\setlength\tabcolsep{4pt}
\resizebox{0.48\textwidth}{!}{
\begin{tabular}{|l|c|c|c|c|c|c|}
  \hline 
  \multirow{2}{*}{Methods} & \multicolumn{2}{c|}{\textbf{FallingDetection-CeleX}}&\multicolumn{2}{c|}{\textbf{DVSGesture}} & \multirow{2}{*}{\textbf{DailyAction}} & \multirow{2}{*}{\textbf{AR}}\\ 
  \cline{2-5} & Fall Detection  & Action Recognition & 10-classes  & 11-classes & & \\ 
  \hline
  EVENT-DRIVEN~\cite{8891738}         &-- & -- &-- & --   & 68.3 & 55.0  \\ 
  SPA~\cite{liu2020effective}         &-- & -- &-- & --   & 76.9 & -- \\
  Truenorth~\cite{amir2017low}        &-- & --& 91.8 & -- &-- & --\\
  SLAYER~\cite{shrestha2018slayer}    &-- &-- & 93.6 & -- &-- &--  \\
  Motion-based SNN~\cite{liu2021event}  & --  & --   & -- &  92.7  & 90.3 & 78.1\\
  PlainNet*~\cite{fang2021deep}         & 89.8 & 64.3 & 92.1 & 91.7 & 92.8 & 66.7 \\
  Spiking ResNet*~\cite{fang2021deep}    & 91.8 & 70.4 & 92.1 & 90.6 & 95.8 & 67.4 \\ 
  SEW ResNet-ADD*~\cite{fang2021deep}    & 93.2 & 82.7 & 97.4 & 97.2 & 98.7 & 83.0 \\ 
  EvT~\cite{sabater2022event}    & -- & -- & 98.5 & 96.2 & -- & -- \\ 
  \hline
  \textit{TS-SNN} (Ours)  & 94.6 & 88.4 & \textbf{98.9} & \textbf{97.6} & \textbf{99.4} & 89.1 \\ 
  \textit{3D-SNN} (Ours)   & \textbf{95.9} & \textbf{90.1} & 98.1 & \textbf{97.6}  & 99.3 & \textbf{94.9} \\ 
  \hline
\end{tabular}
}
\end{center}
\label{tab:All experimental results}
\vspace{-0.7cm} 
\end{table}

\vspace{-0.15cm}
\subsection{Comparison with Existing Literature}
\noindent\textbf{FallingDetection-CeleX Dataset.}
Table~\ref{tab:All experimental results} shows the experimental results on \textit{FallingDetection-CeleX} Dataset.
Our proposed methods achieve state-of-the-art performance in both the `Fall Detection' and `Action Recognition' settings.
Notably, in the `Action Recognition' setting, the proposed \textit{3D-SNN} outperforms the existing method by more than $7\%$. 

\noindent\textbf{DVSGesture Dataset.}
Following the cross-subject protocol suggested in \cite{amir2017low}, we compare \textit{3D-SNN} and \textit{TS-SNN} with state-of-the-art methods.
Since there is an extra category for random movements, Table~\ref{tab:All experimental results} shows the validation accuracy with and without including this extra category (11 and 10 classes classification, respectively). 
Our proposed methods outperform state-of-the-art methods in both settings.

\noindent\textbf{DailyAction Dataset.}
Since there is no official training and validation set, we conduct the experiments based on the 5-fold cross-validation strategy.
As shown in Table~\ref{tab:All experimental results}, our proposed methods also achieve state-of-the-art performance.

\noindent\textbf{AR Dataset.}
We compare our proposed methods with state-of-the-art methods, as shown in Table~\ref{tab:All experimental results}. We can find that our proposed \textit{TS-SNN} and \textit{3D-SNN} outperform the other SNN-based methods by a large margin. 

\vspace{-0.15cm}
\subsection{Ablation Studies}
In this subsection, all ablation studies are conducted on the \textit{FallingDetection-CeleX} Dataset's `Action Recognition' task.


\noindent\textbf{1) Impact of the spatial and temporal convolutional kernel sizes:}
The convolutional kernel size has a direct effect on the learned features, 
here we present the classification accuracy obtained using different spatial and temporal convolutional kernel sizes with \textit{3D-SNN} architectures, as shown in Table~\ref{tab:abla kernel size}.

\begin{table}
\caption{Comparisons of \textit{3D-SNN} with different 3D kernel sizes.
}
\vspace{-0.2cm}
\centering
\resizebox{.28\textwidth}{!}{
\setlength\tabcolsep{4pt}
\begin{tabular}{|c|c|c|c|}
  \hline 
 \multicolumn{2}{|c|}{} & $f_t \times f_w \times f_h$ &   Top-1 Acc. ($\%$)  \\
  \hline
  \multirow{5}{*}{(a)} & \multirow{5}{*}{Temporal}& $1 \times 3 \times 3$       & 88.1 \\
   &&$2 \times 3 \times 3$       & 88.8 \\
   &&$3 \times 3 \times 3$       & \textbf{90.1} \\
   &&$4 \times 3 \times 3$       & 89.8 \\
   &&$5 \times 3 \times 3$       & 88.4 \\  
  \hline
  \multirow{6}{*}{(b)} &\multirow{6}{*}{Spatial}& $3 \times 2 \times 2$       & 87.8 \\
   &&$3 \times 3 \times 3$       & \textbf{90.1} \\
   &&$3 \times 4 \times 4$       & 89.1\\
   &&$3 \times 5 \times 5$       & \textbf{90.1} \\
   &&$3 \times 6 \times 6$       & 89.1 \\
   &&$3 \times 7 \times 7$       & 88.8 \\
  \hline
\end{tabular}
\label{tab:abla kernel size}
}
\vspace{-0.4cm}
\end{table} 

\noindent\textbf{Temporal kernel sizes:} Row (a) in Table~\ref{tab:abla kernel size} shows results of {\textit{3D-SNN}} when varying temporal kernel size. 
We find that the model with a temporal kernel size $3$ performs best among the different temporal kernel sizes. 
Consequently, we choose the temporal kernel size $3$ for all experiments in this paper.

\noindent\textbf{Spatial kernel sizes:} 
Table~\ref{tab:abla kernel size} (b) presents the effects of varying spatial kernel sizes on \textit{3D-SNN} performance. 
We can find that the best classification accuracy is 90.1\%, obtained with models using the kernel size of $3 \times 3 \times 3$ and $3 \times 5 \times 5$. 
Since $5 \times 5$ spatial convolutional kernels need much more computation cost, compared with $3 \times 3$ spatial kernel, 
we choose $3 \times 3 \times 3$ for the best trade-off between performance and efficiency for \textit{3D-SNN} experiments in this paper. 

\begin{table}
\centering
\caption{Comparisons of \textit{TS-SNN} with different numbers of segments and numbers of selected frames.
}
\vspace{-0.2cm}
\setlength\tabcolsep{4pt}
\resizebox{.33\textwidth}{!}{
\begin{tabular}{|c|c|c|c|c|c|c|c|}
  \hline 
  \multirow{2}{*}{Segments (X)} & \multicolumn{6}{c|}{ Selected Frames (K)}\\ 
  \cline{2-7} & 2  & 3 & 4 & 5 & 6 & 7 \\ 
  \hline
   3 &  86.4 & 88.1 & 87.8 & \textbf{88.4} & \textbf{88.4} & 87.4 \\
   \hline
   4 &  86.4 & 87.8 & 88.1 & 88.1 & -- & --\\
   \hline
   6 &  87.4 & 87.8 & --& --& --& --\\
   \hline
   8 &  88.1 & --& --& --& --& --\\
  \hline

\end{tabular}
\label{tab:group and frames}
}
\vspace{-0.6cm}
\end{table}

\noindent\textbf{2) Impact of the Number of Segments and Random Selected Frames:}
As mentioned in Section~\ref{sec: tsn}, for the \textit{TS-SNN}, we divide the event sequence into $X$ segments and randomly select $K$ frames in each segment. In this part, we study the influence of the number of segments ($X$) and the number of selected frames $K$ on the \textit{FallingDetection-CeleX} Dateset.
As shown in Table~\ref{tab:group and frames}, optimal results are obtained with $L=3, K=5$ and $L=3, K=6$.
Therefore, we choose $L=3, K=5$ for the best trade-off between performance and efficiency. (Note that we do not conduct experiments when only one frame is selected or when all frames are selected.)

\vspace{-0.1cm}
\section{Conclusions}
This paper demonstrates the potential of combining SNNs with event-based cameras for event-based HAR.
To address the limitation of SNNs in processing long-range temporal information, we propose two novel frameworks: \textit{TS-SNN} and \textit{3D-SNN}. 
\textit{TS-SNN} segments actions for efficient long-term information extraction, while \textit{3D-SNN} integrates 3D components to enhance temporal information flow.
%
To encourage further research, we introduce the \textit{FallingDetection-CeleX} dataset, collected using a high-resolution CeleX-V event camera. 
%
%
Our frameworks outperform state-of-the-art methods on this new dataset and three existing neuromorphic datasets, demonstrating their effectiveness in handling long-range temporal information for event-based HAR.
%

{\small
\bibliographystyle{ieee_fullname}
\bibliography{egbib}
}

\end{document}